\documentclass{article}

\usepackage[preprint]{neurips_2025}

\usepackage[utf8]{inputenc} 
\usepackage[T1]{fontenc}    
\usepackage{hyperref}       
\usepackage{url}            
\usepackage{booktabs}       
\usepackage{amsfonts}       
\usepackage{nicefrac}       
\usepackage{microtype}      
\usepackage{xcolor}         
\usepackage{amsmath}
\usepackage{multicol}
\usepackage{multirow}
\usepackage{graphicx}
\usepackage{subcaption}
\usepackage{algorithm,algpseudocode,soul,color}
\usepackage{wrapfig}

\definecolor{lightblue}{rgb}{0.29, 0.59, 0.82}

\usepackage{enumitem}
\usepackage{amsthm}
\usepackage{xcolor}

\newtheorem{definition}{Definition}
\usepackage{makecell}
\usepackage{caption}
\title{ATM‑GAD: Adaptive Temporal Motif Graph Anomaly Detection for Financial Transaction Networks}

\author{%
  Zeyue Zhang \\
    School of Statistics\\
    Renmin University of China \\
  \And
  Lin Song \\
  School of Statistics\\
  Renmin University of China \\
  \AND
  Erkang Bao \\
  School of Statistics\\
  Renmin University of China \\
  \And
  Xiaoling Lu \\
  School of Statistics\\
  Renmin University of China \\
  \And
  Xinyue Wang\thanks{Corresponding author.} \\
  School of Statistics\\
  Renmin University of China \\
}

\begin{document}

\maketitle

\begin{abstract}
Financial fraud detection is essential to safeguard billions of dollars, yet the intertwined entities and fast-changing transaction behaviors in modern financial systems routinely defeat conventional machine learning models. Recent graph-based detectors make headway by representing transactions as networks, but they still overlook two fraud hallmarks rooted in time: (1) temporal motifs—recurring, telltale subgraphs that reveal suspicious money flows as they unfold—and (2) account-specific intervals of anomalous activity, when fraud surfaces only in short bursts unique to each entity. To exploit both signals, we introduce \textbf{ATM-GAD}, an adaptive graph neural network that leverages temporal motifs for financial anomaly detection. A \textit{Temporal Motif Extractor} condenses each account’s transaction history into the most informative motifs, preserving both topology and temporal patterns. These motifs are then analyzed by dual-attention blocks: \textit{IntraA} reasons over interactions within a single motif, while \textit{InterA} aggregates evidence across motifs to expose multi-step fraud schemes. In parallel, a differentiable \textit{Adaptive Time-Window Learner} tailors the observation window for every node, allowing the model to focus precisely on the most revealing time slices. Experiments on four real-world datasets show that ATM-GAD consistently outperforms seven strong anomaly-detection baselines, uncovering fraud patterns missed by earlier methods.

\end{abstract}

\section{Introduction}\label{sec:intro}

Financial fraud has escalated alongside the growth of digital finance, causing hundreds of billions of dollars loss each year—more than \$400 billion in the United States alone \cite{bhattacharyya2011data,kirkos2007data}. The sheer volume and velocity of modern transactions make manual auditing infeasible, so automated fraud-detection systems have become indispensable for financial institutions and regulators.

Early 
detection efforts primarily relied on traditional machine learning models such as Random Forests \cite{breiman2001random} and XGBoost \cite{chen2016xgboost}. These models treat each transaction independently and therefore overlook the entity relationships.  
Recent work remedies this by casting transactions as graphs and analyzing higher-order connectivity patterns. Of particular interest are triadic motifs—directed 3-node, 3-edge subgraphs—which naturally encode the canonical ``payer--mule--beneficiary'' chains observed in money-laundering schemes \cite{egressy2024provably, paranjape2017motifs, raphtory2025}. 
Moreover, triads offer an attractive tradeoff between expressiveness and tractability. As the number of candidate motifs grows exponentially with motif size \cite{paranjape2017motifs}, searching for 4–6-node patterns quickly becomes infeasible on large-scale transaction graphs (more analysis in Appendix~\ref{App::Motifs Selection reason}). 



Despite their success, current motif-centric approaches overlook two temporal factors that are critical in practice. First, they treat motifs as static. In reality, financial networks are dynamic, and the precise timing of edges often reveals suspicious money flows (Appendix \ref{APP::Overall Illutration} Figure \ref{fig:insight1}). Second, they impose a single global observation window—typically the full lifespan of the dataset—on every account, ignoring the fact that different entities operate on different timescales (Appendix \ref{APP::Overall Illutration} Figure \ref{fig:insight2}).  Both simplifications blur short-lived yet high-impact fraud signals.

\begin{wrapfigure}{r}{0.45\linewidth}  
    \centering
    \includegraphics[width=\linewidth]{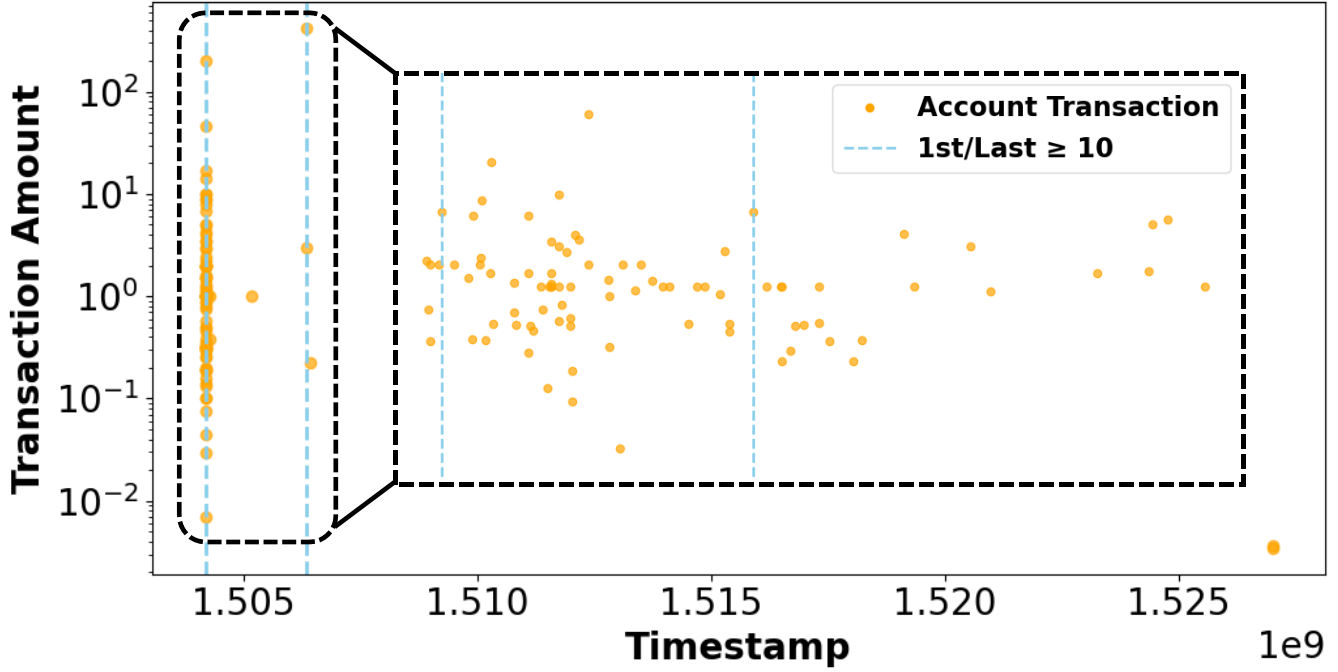}
    \caption{Transactions of fraudulent account $v_0$ over time from the Ethereum dataset\cite{xblockEthereum}. 
    {During its early stage {(dashed box)}, the account initiates a burst of large transfers; afterwards activity recedes to occasional low-value transactions. Even within the burst, the flow evolves from small transfers to a sudden spike of high-value ones and back again. Using the entire time span as a window dilutes the anomalous burst with benign tail activity, while an overly narrow window risks missing the shape of the burst itself. Static triads are likewise insufficient. The very timing and ordering of edges is what distinguishes this laundering episode from legitimate accounts. }}
    \label{fig:ex_illu}
\end{wrapfigure}
Motivated by these limitations and the illustrative example \footnote{Account number 0x44a7ff01f7d38c73530c279e19d31527bdcf8c78 from the Ethereum dataset \cite{xblockEthereum}.} shown in Figure \ref{fig:ex_illu}, we extend the static motif approach to \textit{temporal motifs}. Our approach aims to couple structural and {time-sensitive} information within a GNN to effectively capture the complex fraudulent behaviors. This new setting introduces two challenges. First, unlike static motifs, extracting temporal motifs hinges on the choice of time window. Rather than a single fixed hyper-parameter, the window length must be adapted individually for each node. Second, in the financial transaction networks, each account associates with a unique mix of motif types, and the number of candidate motifs grows rapidly with both window length and graph size, complicating the identification of informative patterns. 

To address the challenges, we propose \textbf{ATM-GAD}, an adaptive graph neural network for financial fraud detection. Our solution integrates three novel components: (1) a \textit{Temporal Motif Extractor} that identifies time-sensitive subgraph patterns critical for fraud detection; (2) dual attention mechanisms (\textit{IntraA} and \textit{InterA}) that process structural information at different granularities; and (3) an \textit{Adaptive Time Window Learner} that optimizes entity-specific observation periods through a differentiable framework. This design enables effective capture of both evolving network structures and account-specific anomalous activities.

The key contributions of our work are as follows:
\begin{itemize}[leftmargin=10pt, itemsep=2pt, topsep=1pt]
    \item 
    We develop a novel approach that extracts time-sensitive motifs within learned node-specific time windows, capturing recurring subgraphs that reveal suspicious money flows while adapting to the short bursts of anomalous activity unique to each entity. 
    
    \item 
    Our framework introduces a dual-level attention mechanism—\textit{IntraA} for reasoning within motifs and \textit{InterA} for aggregating across motif types—that identify complex fraud schemes spanning multiple transaction patterns.
    
    \item 
    Experiments on four real-world datasets demonstrate ATM-GAD consistently outperforms seven SOTA detection baselines, uncovering previously missed fraud patterns.
\end{itemize}

\section{Related Work}\label{sec:related_work}

\paragraph{Local structure-based graph anomaly detection.}
In financial networks, fraudulent users often exhibit distinctive local structural patterns, particularly in collaborative fraud scenarios. Existing methods primarily leverage subgraphs to capture these patterns. ARISE \cite{duan2023arise} targets high-density substructures associated with illegal activities using random walks. Similarly, GRADATE \cite{duan2023graph} and SAMCL \cite{hu2023samcl} employ random-walk-based subgraph sampling with contrastive learning. While these approaches capture certain structural information, they are computationally expensive for large networks and neglect time-varying patterns. In contrast, our ATM-GAD incorporates temporal motifs and adapts to local temporal dynamics, enabling efficient detection of anomalies arising from evolving collaborative fraud patterns.

\paragraph{Motif-based graph anomaly detection.}

Motif-based methods have attracted great attention due to their ability to capture higher-order structural patterns, which are critical for detecting subtle anomalies. For example, MotifCAR \cite{xiao2024motif} uses counterfactual analysis based on static motifs to infer causal relationships, while HO-GAT \cite{huang2021hybrid} combines graph attention mechanisms with motif-based learning to detect anomalies at both the node and subgraph levels. Additionally, MotifGNN \cite{wang2023financial} and MCoGCN \cite{xiang2024mcogcn} integrate motif structures into GNNs, improving the detection of global structural anomalies. 
However, these approaches treat motifs as static entities and do not account for temporal variations or node-specific motif distributions. Our proposed ATM-GAD addresses these limitations by incorporating a Temporal Motif Extractor that dynamically captures time-sensitive motif patterns for each node, and by introducing novel Intra-Attention and Inter-Attention mechanisms that refine both local and cross-motif interactions.


\paragraph{Temporal motif in networks.}
Temporal motifs extend static motifs by incorporating time dimensions, revealing network dynamics. While efficient counting algorithms exist \cite{gao2022scalable, paranjape2017motifs}, they use time windows as fixed hyperparameters. Recent applications include: COFD \cite{hu2023collaborative} using second-order relationships; SLADE \cite{lee2024slade} and approaches by \cite{sarpe2024scalable} demonstrating effectiveness for evolving interactions; and MTM \cite{liu2023using} showing how motif transitions preserve structural information. However, these methods generally employ fixed time windows for extraction, ignoring varying temporal behaviors across nodes. ATM-GAD addresses this limitation through an adaptive window learning mechanism that automatically adjusts each node's extraction window based on local network characteristics.

\section{Preliminaries and Problem Formulation}\label{sec:preliminaries}







\subsection{Financial Transaction Graph}

\begin{definition}[Financial Transaction Graph]\label{Def1} We present a Financial Transaction Graph (FTG) as a directed labeled graph $G = (\mathcal{V}, \mathcal{E}, X, T, Y)$ modeling dynamic financial relationships, where $\mathcal{V}$ denotes financial entities with $|\mathcal{V}| = n$, $\mathcal{E}$ represents directed transaction edges, $X \in \mathbb{R}^{n \times d}$ is the node feature matrix, $T = \{t_v\}_{v \in \mathcal{V}}$ captures activity timestamps, and $Y = \{y_v\}_{v \in \mathcal{V}} $ assigns labels $ y_v \in \{0, 1\} $ indicating non-fraudulent (0) or fraudulent (1) entities. 
\end{definition}

\subsection{Temporal Motifs}
\begin{definition}[$k$-node $l$-edge Motif] 
    A motif $M = (\mathcal{V_M},  \mathcal{E_M})$ is a connected small, recurrent subgraph, where $\lvert \mathcal{V}_M \rvert = k$ and $\lvert \mathcal{E}_M \rvert = l$, such that $\mathcal{V}_M \subseteq \mathcal{V}$ and $\mathcal{E}_M \subseteq \mathcal{E}$.
\end{definition}
While static motifs provide insights into local structure, financial interactions are inherently time-sensitive. We therefore extend the notion of motifs to incorporate temporal information. 
\begin{definition}[$k$-node $l$-edge Temporal Motif]\label{Def3}
    In a dynamic graph $G = (\mathcal{V}, \mathcal{E}, T)$, a temporal motif is defined as a pair $\mathcal{M} = (M, \gamma)$, where $M = (\mathcal{V}_M, \mathcal{E}_M)$ is a static motif extracted from $G$, and $\gamma = \{t_e : e \in \mathcal{E}_M\}$ is the set of timestamps associated with the edges of $M$, such that the temporal span satisfies $t_{\max} - t_{\min} \leq \delta$, for a given time window $\delta$.
\end{definition}
\textbf{Remark.} Definition~\ref{Def3} {emphasizes} two key characteristics overlooked by previous studies \cite{hu2023collaborative, wang2023financial}: (1) The choice of the time window $\delta$ can drastically influence which temporal motif sets are most indicative of fraud \cite{liu2023temporal}; and (2) the temporal motif instances sets differ across nodes in the graph. In this work, we primarily focus on the $3$-node $3$-edge temporal motif for the tradeoff between expressiveness and tractability.

\subsection{Problem Formulation}




\begin{definition}[Temporal Motif–Aware Financial Fraud Account Detection] Given a financial transaction graph \(G = (\mathcal{V}, \mathcal{E}, X, T, Y)\) as defined in Definition \ref{Def1}, let $\delta_v\in(0,\tau_{\max}\ ]$ be a node-specific, learnable time window and $\mathcal M_v(\delta_v)$ be the set of temporal-motif instances associated with node $v$ as defined in Definition \ref{Def3}. We seek to solve $\max_{\eta} \sum_{v \in \mathcal{V}} \mathbb{I}\left( f_{\eta}(G,\ \mathcal M_v(\delta_v) )_v = y_v \right)$, where \(y_v \in \{0,1\}\) is the ground-truth label for node \(v\), and \(\mathbb{I}(\cdot)\) is the indicator function.
In implementation, we use cross-entropy loss to approximate this objective. \end{definition}

\section{Methodology}\label{sec:methods}
\begin{figure*}
    \centering
    \includegraphics[width=1.0\linewidth]{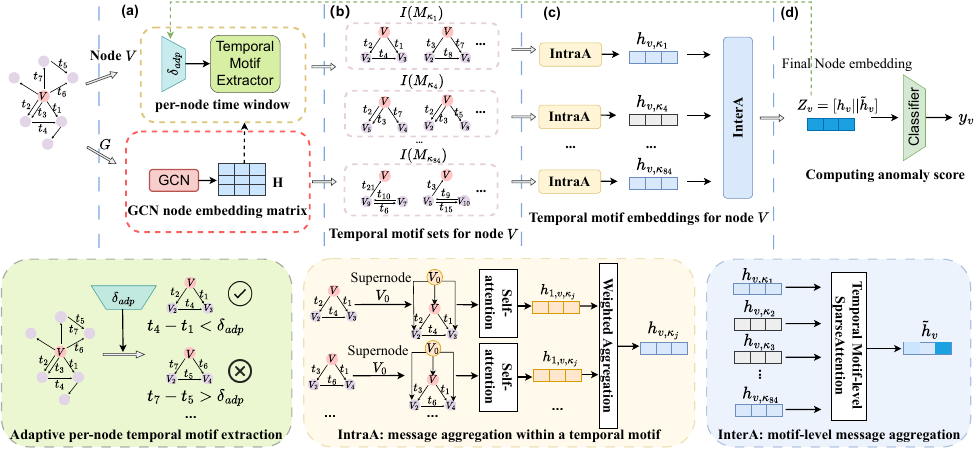}
    \caption{The Overview of ATM-GAD. (a) Node embeddings are computed via GCN\cite{kipf2016semi}, along with a per-node adaptive time window. (b) Node-specific temporal motifs are extracted using the adaptive time window. (c) Intra- and Inter-Attention modules aggregate motif-level information to produce refined motif embeddings. (d) An anomaly score is then calculated based on concordanced node embedding for fraud detection.
    }
    \label{fig:framework}
\end{figure*}

In this section, we present the \textbf{ATM-GAD} framework (Figure~\ref{fig:framework} and Appendix~\ref{APP::pseudocode} Algorithm~\ref{alg:modified}), which provides a flexible architecture where various graph neural networks (e.g., GCN, GAT) can serve as the backbone. Our implementation uses GCN (Appendix~\ref{App::GCN}) as the backbone. The fraud detection process consists of the following steps:
(1) Initialize node representations using GCN to encode account features into node embeddings;
(2) For each node in the graph, learn an adaptive time window and extract node-specific temporal motifs within this window;
(3) Apply Intra- and Inter-Motif Aggregation mechanisms to aggregate information within each motif and across various motif types, producing refined node representations;
(4) Generate fraud predictions based on the final node embeddings and update the parameters.

\subsection{Adaptive Per-node Temporal Motif Extraction}\label{Sec::Adaptive_Temporal_Motif_Extraction}
Traditional methods that ignore temporal dynamics or use global motif extraction fail to capture node-specific transaction behaviors. Our proposed approach extracts temporal motifs adaptively for each node, combining two key innovations: (1) adaptive time window learning and (2) node-specific temporal motif extraction.

\noindent\textbf{Adaptive time window learning.$\quad$}
For each node $v_i$ in the graph, we first learn an adaptive time window $\delta_{v_i}$ through a neural network $f_\theta(\cdot)$ that maps the node embeddings to the optimal time bound. To ensure the learned time window is meaningful, as neural networks typically output unconstrained values that may not serve as valid time windows, we employ a reparameterization technique with carefully designed bounds:
\begin{equation}\label{eq:deltaada}
    \delta_{v_i} = \tau_\texttt{max} \cdot \sigma\left( f_{\theta}(\mathbf{z}_{v_i}) \right)
\end{equation}
where $\mathbf{z}_{v_i}$ denotes the final node embeddings, $\tau_\texttt{max}$ is the latest timestamp in the graph, and $\sigma(\cdot)$ is the sigmoid function. This design is critical as it guarantees that: (1) $\delta_{v_i}$ is always positive, which is essential for a valid time window; (2) $\delta_{v_i}$ is upper-bounded by $\tau_\texttt{max}$, preventing the extraction of irrelevant historical patterns; and (3) the continuous nature of the sigmoid function ensures smooth gradient flow during optimization; and (4) because $\delta_{v_i}$ is used in weighted pooling (Equation~\ref{eq:intra-weight}), the learning process remains fully differentiable, allowing $\delta_{v_i}$ to be updated via gradient-based methods (details in Appendix~\ref{APP::Adaptive Time Window Update}). This contrasts with previous work \cite{chen2023motif, wang2023financial} that treats $\delta$ as a fixed hyperparameter, despite its crucial impact on fraud detection performance \cite{hu2023collaborative, xie2021temporal}.

\noindent\textbf{Node-specific temporal motif extraction.$\quad$}
Given time window $[t_{v_i}, t_{v_i} + \delta_{v_i}]$, where $t_{v_i}$ is $v_i$'s earliest timestamp, we extract all possible \emph{3-node 3-edge} temporal motifs involving $v_i$. Let $\{\kappa_1, \kappa_2, \dots, \kappa_{\text{Max}}\}$ be the complete set of motif \emph{types}, each representing a distinct configuration (with $\texttt{Max}=84$ in our setting, see Appendix~\ref{App::MotifTypesDescription}). The temporal motif set for $v_i$ is $\mathcal{M}(v_i) = \{\mathcal{M}_{i_1}, \mathcal{M}_{i_2}, \dots, \mathcal{M}_{i_m}\}$, where $\{i_1, \cdots, i_m \} \subseteq \texttt{permutation}\left(\{\kappa_1, \cdots, \kappa_{\text{Max}}\}\right)$.

A single temporal motif can have multiple instances (e.g., $\{v_i, v_2, v_3 \}$ and $\{v_i, v_5, v_7\}$ could both be instances of $\mathcal{M}_1$). Let $I_{v}(\mathcal{M})$ denote all instances of $\mathcal{M}$ including node $v$. Thus, $I\left(\mathcal{M}(v_i)\right) = \bigl\{ I_{v_i}(\mathcal{M}_1), I_{v_i}(\mathcal{M}_2), \dots, I_{v_i}(\mathcal{M}_m) \bigr\}$ represents the final collection of temporal motif instances for $v_i$. These instances preserve both \emph{local structural patterns} and \emph{temporal edge ordering}, providing a rich foundation for anomaly detection.

\subsection{Intra- and Inter-Motif Aggregation}

Financial anomalies manifest as distinct behaviors diverging from normal transaction patterns, such as tightly orchestrated transfers over brief periods. In temporal motifs, each node has a unique mixture of motif types and varying instance counts per type. Furthermore, in large-scale financial networks, possible motif numbers grow rapidly with increasing time windows and graph size. To address these challenges, we propose a two-level attention mechanism: \textbf{Intra-A}ttention to capture relationships \emph{within} each motif, and \textbf{Inter-A}ttention to aggregate information \emph{across} different motif types.

\paragraph{Intra-attention mechanism.}
To aggregate node information within motif instances, we adopt the \emph{supernode} technique inspired by \cite{yuan2023motif}. Consider a \emph{3-node} motif instance of $I_{v_i}(\mathcal{M})$ involving nodes  $\{v_{i}, v_{i2}, v_{i3}\}$, we introduce a conceptual node $v_0$, forming an augmented motif instance $\overline{M}$. To reduce model complexity, we use type-specific supernode embedding to aggregate information across temporal motif instances of a given type.

For motif instance $\mathcal{M}^u$ involving nodes $\{v_{i}, v_{i2}, v_{i3}\}$, we initialize supernode embedding, $\mathbf{h}_{0, (u, v_i, \kappa_j)}$, and assemble 
\begin{equation}
    H_{\overline{\mathcal{M}}^u}= \left[\mathbf{h}_{0, (u, v_i, \kappa_j)}, \mathbf{h}_{v_i},\, \mathbf{h}_{v_{i2}},\, \mathbf{h}_{v_{i3}}\right], 
\end{equation}
where $\mathbf{h}_{v_i},\, \mathbf{h}_{v_{i2}},\, \mathbf{h}_{v_{i3}}$ are the node embeddings obtained via the GCN, and $\mathbf{h}_{0, (u, v_i, \kappa_j)}$ is the supernode embedding corresponding to $\overline{\mathcal{M}^u}$. 

To obtain the instance representation, we then apply the self-attention mechanism \cite{vaswani2017attention}. In detail, the attention score for each motif is defined as 
$s_{\texttt{intra}}\left(\mathbf{h}_v\right)=\text{tanh}\left( \mathbf{w}^v_\texttt{intra}\mathbf{h}_v\right), v \in \mathcal{V}_{\overline{\mathcal{M}}^u}$,
where $\mathbf{w}^{v}_\texttt{intra}$ denotes the learnable parameters that map the node embedding into a score.

Then, the motif embedding of $\mathcal{M}^u$, $\mathbf{h}_{u, v_i, \kappa_j}$, is obtained with the updated supernode embedding.
\begin{equation}
\mathbf{h}_{u, v_i, \kappa_j} = \sum_{v \in \mathcal{V}_{\overline{\mathcal{M}}^u}} \alpha_{v} \mathbf{h}_v, 
\end{equation}
where $\alpha_{v} = \frac{\exp(s_{\texttt{intra}}\left(\mathbf{h_v}\right)}{\sum_{v \in \overline{\mathcal{M}}^u} \exp(s_{\texttt{intra}}\left(\mathbf{h_v}\right)}$ are the attention weights.

Each motif type $\kappa_j$ may have multiple instances $I_{v_i}(\mathcal{M}{\kappa_j})$ that include node $v_i$. We then assign a weight to each instance $\mathcal{M}^u \in I_{v_i}(\mathcal{M}{\kappa_j})$ to emphasize more recent temporal patterns:
\begin{equation}\label{eq:intra-weight}
    w_{u, v_i, \kappa_j} = \sigma\left( \delta_{v_i} - (\tau^u_{max} - t_{v_i})\right), 
\end{equation}
where $ \sigma(\cdot)$ denotes the sigmoid function, $\tau^u_{max}$ denotes the latest timestamp among the edges in $\mathcal{M}^u$, and $t_{v_i}$ denotes the earliest timestamp associated with node $v_i$.
This weighting mechanism has two key advantages: Firstly, it 
naturally integrates $\delta_{v_i}$ (the learnable time window for $v_i$) into the calculation, allowing gradients to flow back and update $\delta_{v_i}$ accordingly. 
Secondly, it naturally filters out temporally distant patterns (when $\tau^u_{max} - t_{v_i} \gg \delta_{v_i}$), emphasizing more recent patterns that generally carry more meaningful information about the current node behavior.

The \emph{type-specific} motif embedding for $\kappa_j$ with respect to node $v_i$ is then:
\begin{equation}
\mathbf{h}_{v_i, \kappa_j} = \frac{\sum_{u} w_{u, v_i, \kappa_j} \cdot \mathbf{h}_{u, v_i, \kappa_j}}{\sum_{u} w_{u, v_i, \kappa_j}}.
\end{equation}

\paragraph{Inter-attention mechanism.}
To integrate the embeddings of all 84 motif types, we introduce a \textbf{motif-level attention} mechanism.
In particular, we apply a sparse attention mechanism \cite{child2019generating} that assigns varying levels of importance to each motif type, which enhances computational efficiency while maintaining expressive power. 
The attention score is defined as:
$s_{\texttt{inter}}\left(\mathbf{h}_{\kappa_j} \right)=\text{tanh}\left( \mathbf{w}^{\kappa_j}_\texttt{inter}\mathbf{h}_{\kappa_j}\right), $
where $\kappa_j \in \{\kappa_1, \kappa_2, \dots, \kappa_{\text{Max}}\}$, $\mathbf{w}^{\kappa_j}_\texttt{inter}$ denotes the learnable parameters associated with each motif type.

We then apply \texttt{SparseMAX} function to derive sparse attention weights:
\begin{equation}
    \beta_{v_i,\kappa_j}=\texttt{SparseMAX}\left(\left[s_{\texttt{inter}}(\mathbf{h}_{v_i, \kappa_1}), \cdots, s_{\texttt{inter}}(\mathbf{h}_{v_i, \kappa_\texttt{Max}}) \right] \right).
\end{equation}
The temporal motif embedding for node $v_i$ is  simply the sum over all the motif-level representations:
\begin{equation}
\tilde{\mathbf{h}}_{v_i} = \sum_{\kappa_j} \beta_{v, \kappa_j} \mathbf{h}_{v_i, \kappa_j}.
\end{equation}

\paragraph{Final node embedding.}
To incorporate both global node embeddings (from the backbone GNN) and motif-level information, we concatenate $\mathbf{h}_{v_i}$ with $\tilde{{\mathbf{h}}}_{v_i}$, to form the final node embedding $\mathbf{z}_{v_i}$: 
$\mathbf{z}_{v_i} = \left[ \mathbf{h}_{v_i} \, \| \, \tilde{{\mathbf{h}}}_{v_i} \right].$
This design provides flexibility as various graph neural networks (e.g., GCN\cite{kipf2016semi} and GAT\cite{velivckovic2017graph}) can serve as the backbone to generate the initial node embeddings $\mathbf{h}_{v_i}$, while the temporal motif embedding $\tilde{{\mathbf{h}}}_{v_i}$ captures the critical temporal and structural patterns specific to each node. 
{This concatenation strategy preserves both global structural information and local temporal dynamics, allowing the model to leverage these complementary aspects for enhanced fraud detection.}

\subsection{Model Training and Complexity}
\noindent\textbf{Classification. $\quad$} Once obtained the final node representations, $ \mathbf{z}_{v_i} $, we pass them through a classifier to produce the prediction score, $ \hat{y}_{v_i} $. In the implementation, we adopt a two-layer MLP followed by a sigmoid activation function:
    $\hat{y}_{v_i} = \sigma\left( f_\eta(\mathbf{z}_{v_i})\right),$
where $f_\eta(\cdot):\mathbb{R}^{2d}\rightarrow \mathbb{R}$ is the classifier, parameterized by $\eta$, and $ \sigma(\cdot):\mathbb{R}\rightarrow (0,1) $ denotes the sigmoid activation function.

The training objective is the standard binary cross-entropy loss:
\begin{center}
    $\mathcal{L} = \frac{1}{N} \sum_{v_i \in \mathcal{V}} \left( - y_{v_i} \log \hat{y}_{v_i} - (1 - y_{v_i}) \log (1 - \hat{y}_{v_i}) \right), 
$
\end{center}
where $ y_{v_i}\in \{0,1\} $ denotes the ground-truth label for the node $v_i$. 

\noindent\textbf{Parameters update.$\quad$} All model parameters—from the GCN, Temporal Motif Extractor, attention modules, and MLP—are jointly optimized via backpropagation.

\noindent\textbf{Complexity analysis.$\quad$} The total complexity of ATM-GAD is dominated by \( \mathcal{O}(|\mathcal{V}| \bar{d}^2) \), where \( |\mathcal{V}| \) is the number of nodes and \( \bar{d} \) is the average degree of the graph (which is relatively small in real-world sparse networks \cite{akcora2017blockchain, chen2020understanding}, see Appendix~\ref{App::DatasetsDescription}), more details in Appendix~\ref{APP::Complexity Analysis}.
\section{Experiments}\label{sec:results}

In this section, we demonstrate the effectiveness of our method through extensive experiments on four real-world datasets, each with varying levels of network complexity. We begin by detailing the experimental settings, followed by an in-depth presentation and analysis of our results.

\subsection{Experimental Settings}
\noindent\textit{Dataset.} We include four datasets widely used in financial graph anomaly detection research, which are \underline{\smash{ETH}}~\cite{xblockEthereum}, \underline{\smash{Elliptic++}}~\cite{elmougy2023demystifying}, \underline{\smash{Bitcoin Alpha}}~\cite{kumar2016edge}, and \underline{\smash{Bitcoin Otc}}~\cite{kumar2018rev2}. 
To assess performance and efficiency under different network scales and time horizons, we further create smaller or larger subgraphs, yielding a total of eight dataset variants.
We then perform three independent train–test splits to prevent information leakage. 
The detailed description of each dataset and the sampling strategies is presented in Appendix \ref{App::DatasetsDescription}.



\begin{table*}[]
\centering
\caption{Node classification performance comparison. The best results are in \textbf{bold} and the second-best are \underline{\smash{underlined}}. The numerical suffix following the ‘-’ in each dataset name corresponds to its size. Complete results with standard deviations are presented in Appendix~\ref{App:Performance Comparison with SOTA}, Table~\ref{apptable::SupPerformanceComparisonwithSOTA}.}
\resizebox{0.99\textwidth}{!}{
\renewcommand{\arraystretch}{1.15}
\begin{tabular}{ccccccccccccccccc}
 \hline
Dataset       & \multicolumn{2}{c}{\textbf{ATM-GAD}} & \multicolumn{2}{c}{GraphSage~\cite{hamilton2017inductive}} & \multicolumn{2}{c}{ARISE~\cite{duan2023arise}} & \multicolumn{2}{c}{HO-GAT~\cite{huang2021hybrid}} & \multicolumn{2}{c}{MotifGNN~\cite{wang2023financial}} & \multicolumn{2}{c}{COFD~\cite{hu2023collaborative}} & \multicolumn{2}{c}{Random Forest~\cite{breiman2001random}}          & \multicolumn{2}{c}{XGBoos~\cite{chen2016xgboost}} \\  \hline
Metric        & AUC                       & AUPRC                     & AUC           & AUPRC         & AUC         & AUPRC       & AUC             & AUPRC    & AUC           & AUPRC        & AUC      & AUPRC         & AUC            & AUPRC          & AUC      & AUPRC            \\ \hline
ETH-200      & \textbf{0.911} & \textbf{0.739} & 0.621 & 0.512 & 0.711 & 0.609 & 0.781 & 0.464 & \underline{0.875} & 0.596 
& 0.768 & \underline{0.633} & 0.821 & 0.562 & 0.608 & 0.279 \\
ETH-1000      & \textbf{0.961} & \textbf{0.844} & 0.910 & 0.699 & 0.749 & 0.445 & 0.922 & 0.614 & \underline{0.946} & 0.716 
& 0.775 & \underline{0.723} & 0.864 & 0.503 & 0.709 & 0.412 \\
ETH-1200     & \textbf{0.968} & \textbf{0.872} & 0.763 & 0.538 & 0.785 & 0.544 & 0.910 & 0.629 & \underline{0.932} & 0.724 
& 0.860 & \underline{0.757} & 0.855 & 0.497 & 0.802 & 0.359 \\
Elli-500     & \textbf{0.944} & \textbf{0.911} & 0.601 & 0.528 & 0.817 & 0.761 & 0.703 & 0.691 & 0.880 & \underline{0.887} 
& 0.853 & 0.875 & \underline{0.930} & 0.816 & 0.858 & 0.820 \\
Elli-1000    & \underline{0.922} & \textbf{0.796} & 0.740 & 0.363 & 0.805 & 0.625 & 0.791 & 0.626 & 0.907 & 0.644 
& 0.875 & 0.735 & \textbf{0.935} & 0.643 & 0.569 &  \underline{0.768} \\
Elli-3000    & \underline{0.994} & \textbf{0.997} & 0.500 & 0.541 & 0.852 & 0.788 & 0.712 & 0.725 & 0.982 & 0.986 
& 0.809 & 0.809 & \textbf{0.995} & \underline{0.996} & 0.949 & 0.906 \\
 Bitcoin Alpha & \underline{0.762} & \textbf{0.618} & 0.594 & 0.396 & 0.663 & 0.539 & \textbf{0.768} & 0.552 & 0.739 & \underline{0.607} & 0.699 & 0.555 & 0.702 & 0.437 & 0.670 &0.479 \\
 Bitcoin Otc   & \underline{0.709} & \textbf{0.566} & 0.568 & 0.430 & 0.611 & 0.483 & 0.621 & 0.429 & 0.705 & 0.521 & 0.619 & 0.538 & \textbf{0.714} & \underline{0.550} & 0.620 &0.520 \\  \hline
1st count     & \multicolumn{2}{c}{\textbf{12}}                       & \multicolumn{2}{c}{0}         & \multicolumn{2}{c}{0}     & \multicolumn{2}{c}{1}      & \multicolumn{2}{c}{0}        & \multicolumn{2}{c}{0}    & \multicolumn{2}{c}{{\underline{3}}}     & \multicolumn{2}{c}{0}    \\  \hline
\end{tabular}%
}
\label{table::PerformanceComparisonwithSOTA}
\end{table*}
\noindent\textit{Baseline.} We extensively compare ATM-GAD with the state-of-the-art GAD models, including GNN-based model: \underline{\smash {GraphSAGE}}~\cite{hamilton2017inductive}, subgraph-based model: \underline{\smash ARISE}~\cite{duan2023arise}, motif-based models: \underline{\smash HO-GAT}~\cite{huang2021hybrid}, \underline{\smash MotifGNN}~\cite{wang2023financial}, and temporal graph-based: \underline{\smash COFD}~\cite{hu2023collaborative}. We also include two tree-based methods: \underline{Random Forest} \cite{breiman2001random} and \underline{XGBoost} \cite{chen2016xgboost}.

\noindent\textit{Evaluation Metric.} Following standard practices in financial graph anomaly detection \cite{duan2023arise, wang2023financial}, we measure performance using \underline{AUC} (Area Under the ROC Curve) and \underline{AUPRC} (Area Under the Precision-Recall Curve). Higher values in both metrics signify superior classification performance. For completeness, we also evaluate accuracy and present those results in Appendix~\ref{App:Performance Comparison with SOTA}.


\noindent\textit{Experiment Details.}  
The experiment is implemented using PyTorch 1.8.1 \cite{paszke2019pytorch}, with a GCN of 2-4 layers and dimension sizes \{16, 32, 64\}. 
Further implementation details, including dataset-specific parameters, can be found in Appendix~\ref{App::Experiment Details}. We reproduce all baseline methods using either their official implementations or closely aligned open-source releases. 
We report the key performance results averaged over three splits.

\subsection{Main Results}
\subsubsection{Effectiveness of ATM-GAD}
We present the comparison results in Table \ref{table::PerformanceComparisonwithSOTA}. Overall, \textbf{ATM-GAD} achieves state-of-the-art performance on all settings in terms of \emph{AUPRC}. Several key observations emerge from these results. 
{First}, graph topology significantly enhances fraud detection capabilities. While Random Forest and XGBoost perform well in certain cases (e.g., \textit{Elli-1000}), motif-based models such as ATM-GAD consistently perform well across various settings. This suggests that incorporating graph topology by capturing complex transactional relationships provides additional benefits in detecting fraudulent transactions. 
{Second}, local topology structure proves essential for graph anomaly detection, as evidenced by ATM-GAD, along with other motif-based models (MotifGNN and HO-GAT) and the subgraph-based approach (ARISE), consistently outperforming GraphSAGE. For instance, on \textit{ETH-200}, ATM-GAD and ARISE achieve AUC improvements of 46.70\% and 14.49\%, respectively, over GraphSAGE. 
{Third}, predefined subgraph patterns demonstrate clear advantages over random sampling approaches. When comparing motif-based methods (ATM-GAD, MotifGNN, HO-GAT) with ARISE, all three motif-based approaches outperform ARISE across most datasets, likely because random walk-based subgraph extraction in ARISE introduces noise that reduces detection accuracy. This illustrates how predefined subgraph patterns (motifs) more effectively capture meaningful structures for fraud detection. 
{Fourth}, node-level motif distribution modeling emerges as a crucial factor in detection performance. Unlike MotifGNN and HO-GAT, which focus primarily on motif-based graphs, ATM-GAD models each node's motif distribution individually. This node-level motif learning, combined with a carefully designed attention mechanism, enables ATM-GAD to achieve superior performance by dynamically adapting to localized graph structures (detailed analysis in Sections \ref{sec:res_motif} and \ref{sec:res_delta}). 
{Finally}, adaptive temporal windows prove superior at capturing time-sensitive behaviors. The importance of temporal information is evident in COFD's performance relative to GraphSAGE—on \textit{ETH-200}, COFD achieves an AUC improvement of 23.67\% over GraphSAGE, underscoring the impact of temporal modeling. However, ATM-GAD surpasses COFD by an additional 23.03\% (AUC improvement), demonstrating that our adaptive temporal motif modeling strategy more effectively captures time-sensitive fraud patterns than fixed temporal windows.

\begin{figure}[!h]
    \centering
    \begin{subfigure}[t]{0.68\linewidth}
        \includegraphics[width=\linewidth]{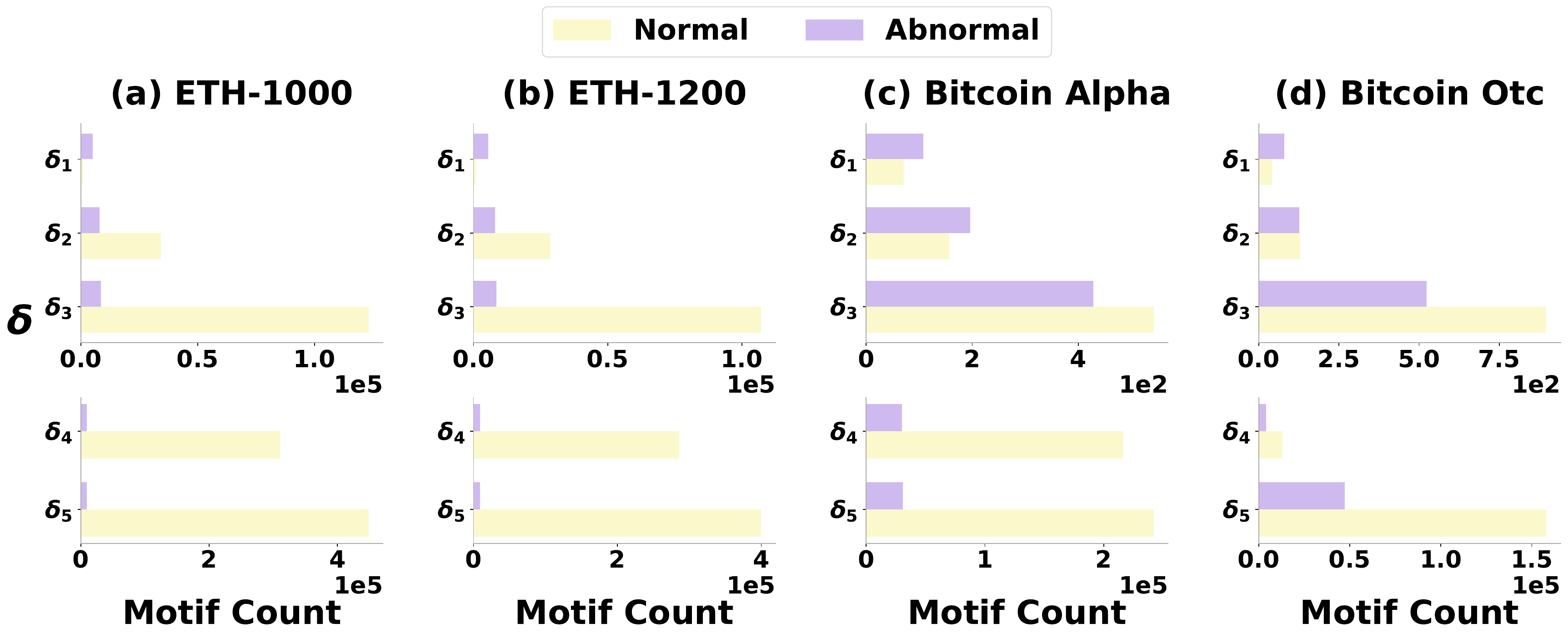}
        \caption{}
        \label{fig::motifdistribution}
    \end{subfigure}
    \hfill
    \begin{subfigure}[t]{0.26\linewidth}
        \includegraphics[width=\linewidth]{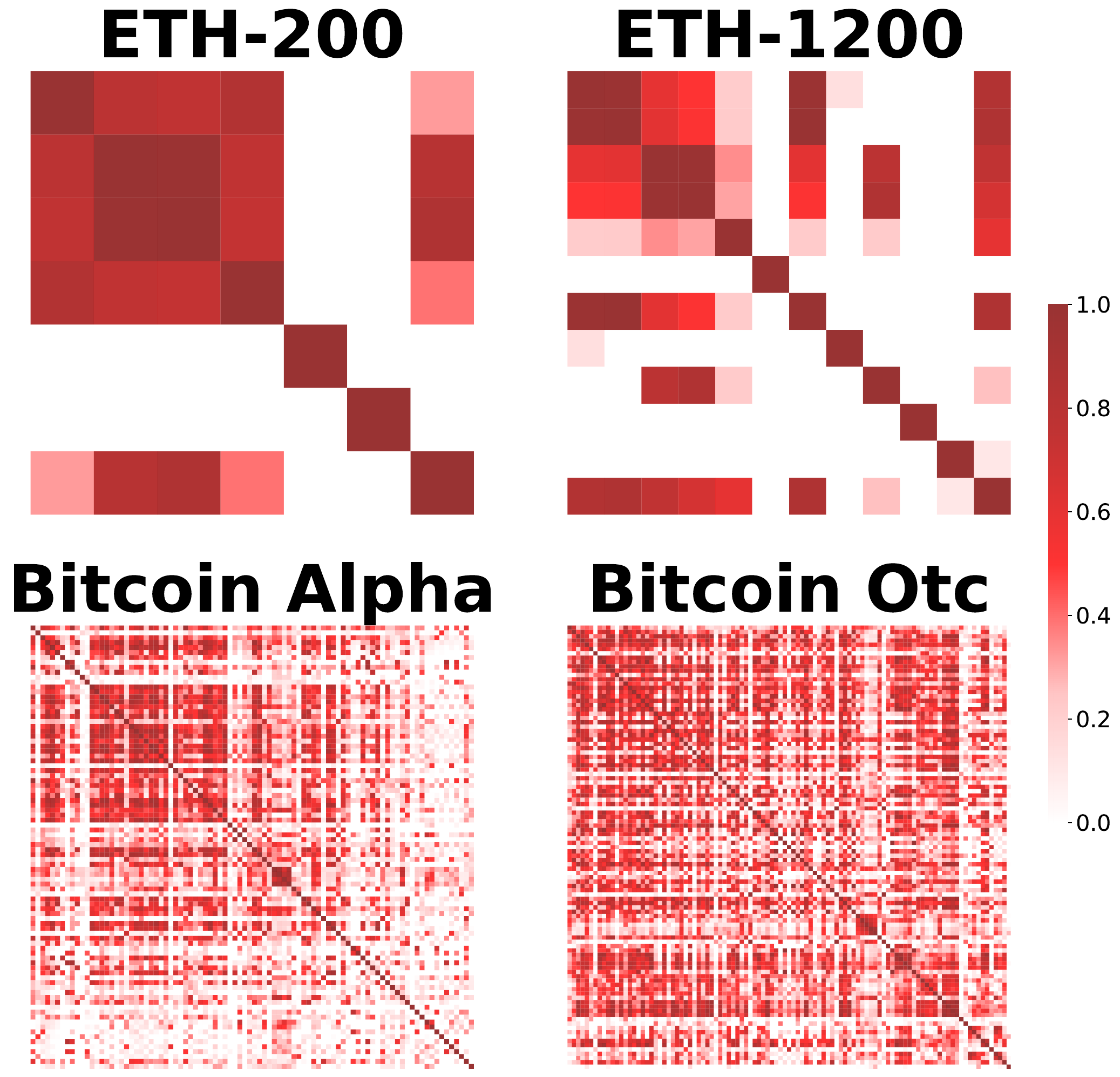}
        \caption{}
        \label{fig::heatmap}
    \end{subfigure}
    \caption{(a) Distribution of temporal motif counts at varying time windows $\delta$ values; (b) Heatmap of cross-correlation coefficients of temporal motif counts among anomalies.
    }
    \label{fig:Motif Distribution}
\end{figure}
\begin{table}[!h]
\centering
\caption{Performance comparison of adaptive ($\delta_\mathrm{ada}$) and fixed ($\delta_\mathrm{fixed}$) time window.
$\tau$ denotes the temporal scope of graph $G$, i.e., the subgraph $G_\tau$ contains all nodes and edges observed in $[0, \tau]$.
}
\resizebox{0.70\linewidth}{!}{%
\begin{tabular}{c|cc|cc|cc|cc} \hline
  Dataset& \multicolumn{4}{c|}{ETH-1200}& \multicolumn{4}{c}{Elli-500}\\
\hline
 \begin{tabular}[c]{@{}c@{}}$\tau$ \end{tabular} & \multicolumn{2}{c|}{$\delta_\mathrm{ada}$} & \multicolumn{2}{c|}{$\delta_\mathrm{fixed}$} & \multicolumn{2}{c|}{$\delta_\mathrm{ada}$} & \multicolumn{2}{c}{$\delta_\mathrm{fixed}$}\\  Metric & AUC & AUPRC & AUC & AUPRC  & AUC &AUPRC  & AUC &AUPRC  \\ \hline
 $\tau_{tiny}$ & \textbf{0.946} & \textbf{0.865} & 0.939 & \textbf{0.865}  & \textbf{0.932} &\textbf{0.933} 
 & 0.872 &0.902  
\\
 $\tau_\mathrm{small}$ & \textbf{0.952} & \textbf{0.867} & 0.943 & 0.861  & \textbf{0.914}  &\textbf{ 0.924}  
 & 0.872  &0.903   
\\
 $\tau_\mathrm{med}$ & \textbf{0.955} & \textbf{0.872} & 0.947 & 0.854  & \textbf{0.926}  &\textbf{0.929}  
 & 0.873  &0.902   
\\
 $\tau_\mathrm{large}$ & \textbf{0.958} & \textbf{0.868} & 0.950 & 0.837  & \textbf{0.930 } &\textbf{0.929}  
 & 0.874  &0.904   
\\
 $\tau_\mathrm{max}$ & \textbf{0.962} & \textbf{0.869} & 0.955 & 0.830  & \textbf{0.925 } &\textbf{0.930 }  & 0.874  &0.904   \\\hline
\end{tabular}%
}
\label{table::deltaavss}
\end{table}
\subsubsection{Effectiveness of Node-level Motif Distribution Modeling}\label{sec:res_motif}
To explore how node-level temporal motif modeling contributes to ATM-GAD’s performance, we first analyze how motifs distribute across normal and anomalous nodes. Figure \ref{fig::motifdistribution} and Appendix \ref{App::MotifAnalysis} show the motif count distributions of normal nodes and anomalies, under different datasets and settings (i.e., different $\delta$ values). 
The results illustrate that anomalous nodes often exhibit higher motif counts in shorter windows compared to normal nodes.
(e.g. $\delta_1$ and $\delta_2$ on \textit{Bitcoin Alpha} and \textit{Bitcoin Otc}). As $\delta$ increases, we observe that normal nodes accumulate motifs at a faster rate than anomalies, implying that \emph{fraudulent users tend to transact intensively over short periods} while legitimate users spread their interactions over longer periods. Additionally, Figure~\ref{fig::heatmap} illustrates the cross-correlation of motif counts among anomalies, revealing distinct patterns of association rather than uniform dominance by any single motif. More comprehensive results are presented in Appendix~\ref{App::heatmap}. This observation further supports our node-specific motif extraction strategy, which accommodates the heterogeneous behaviors of fraudulent.

\begin{wrapfigure}{r}{0.44\linewidth}  
    \centering
    \includegraphics[width=\linewidth]{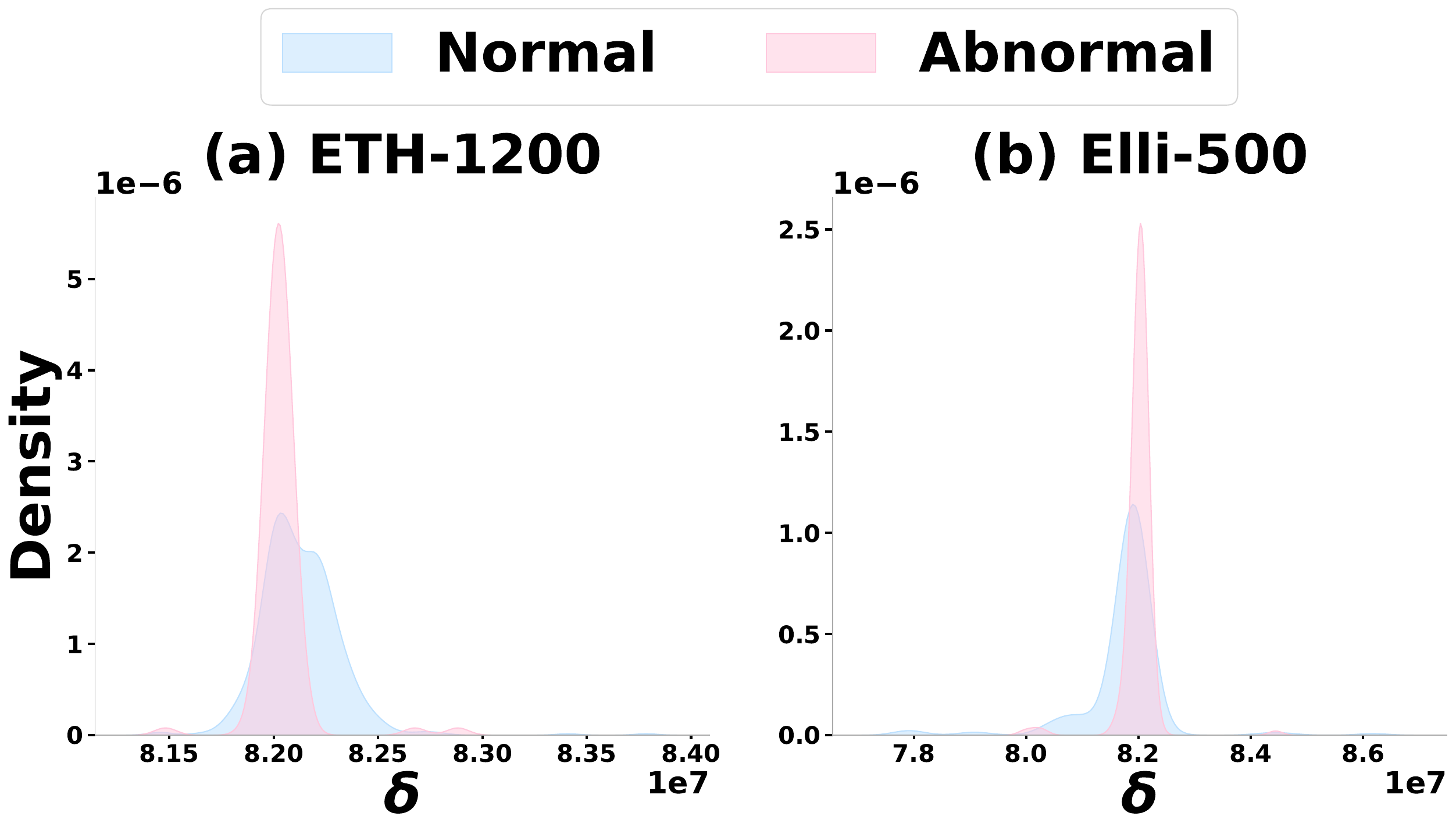}
    \caption{Distribution of the learned adaptive time window  $\delta_\mathrm{ada}$.
    }
    \label{fig::deltadistribution}
\end{wrapfigure}
\subsubsection{Effectiveness of Adaptive Temporal Motif Extraction}\label{sec:res_delta}
We make an in-depth analysis of the proposed adaptive window selection when computing temporal motifs for each node. Firstly, we compare our learnable window $\delta_{\mathrm{ada}}$ with a conventional fixed window $\delta_{\mathrm{fixed}}$. Table~\ref{table::deltaavss} and Appendix \ref{App::Adaptive versus Fixed} report the detection results at different temporal scopes. Our adaptive approach effectively surpasses the fixed window approach consistently. Notably, larger fixed values do not necessarily yield better performance, whereas our adaptive method maintains strong performance. 
For instance, on \textit{ETH-1200}, $\delta_\mathrm{fixed}$ achieves its peak AUPRC (0.865) at $\delta_\mathrm{fixed}=\tau_{\mathrm{tiny}}$. 
Moreover, when $\delta_\mathrm{adp}$ equals $\delta_\mathrm{fixed}$, the adaptive approach consistently yields superior AUC and AUPRC metrics.
However, on \textit{ETH-200}, $\delta_{adp}$ exhibits suboptimal performance, which can be attributed to the dataset's limited size. Figure \ref{fig::deltadistribution} and Appendix \ref{App::deltadistribution} compare $\delta_{\mathrm{ada}}$ distributions between normal and anomalous nodes. Legitimate activities display broader, more uniform $\delta$ distributions, indicating more diverse temporal patterns in normal financial behavior. In contrast, anomalous activities consistently show more concentrated $\delta$ distributions across datasets, though the specific values vary (smaller in \textit{ETH-1200}, slightly larger in \textit{Elli-500}). These consistent observations highlight the dataset-specific nature of $\delta$ values and the need for tailored models. These findings align with the motif accumulation patterns shown earlier, reinforcing the necessity of adaptive temporal modeling.

\begin{table*}[!t]
\centering
\caption{Ablation studies of ATM-GAD on 
 3 ETH datasets. The best results are in \textbf{bold} and the second-best are \underline{\smash{underlined}}. 
 }\label{table::Abaltione}
\captionsetup{skip=5pt}
\resizebox{0.90\textwidth}{!}{
\renewcommand{\arraystretch}{1.15}
\begin{tabular}{c|cc|cc|cc}
\hline
Dataset & \multicolumn{2}{c|}{ETH-200} & \multicolumn{2}{c|}{ETH-1000} & \multicolumn{2}{c}{ETH-1200} \\ \cline{2-7} 
Metric & AUC & AUPRC & AUC & AUPRC & AUC & AUPRC \\ \hline
Only GCN & 0.874(--) & 0.863(--) & 0.930(--) & 0.723(--) & 0.940(--) & 0.840(--) \\
+TM+$\delta_{fixed}$ & 0.894(+2.25\%) & 0.868(+0.57\%) & {\ul{0.946(+1.75\%)}} & 0.733(+1.38\%) & 0.955(+1.59\%) & 0.838(-0.23\%) \\
+TM+$\delta_{ada}$ & 0.892(+2.03\%) & 0.872(+1.08\%) & 0.945(+1.60\%) & {\ul{ 0.745(+3.11\%)}} & 0.946(+0.62\%) & 0.856(+1.92\%) \\
+TM+$\delta_{ada}$+IntraA & {\ul{0.907(+3.83\%)}} & {\ul{0.881(+2.11\%)}} & 0.938(+0.91\%) & 0.714(-1.29\%) & \textbf{0.964(+2.61\%)} & \textbf{0.892(+6.18\%)} \\
+TM+$\delta_{ada}$+InterA & 0.899(+2.93\%) & \textbf{0.886(+2.74\%)} & \textbf{0.953(+2.52\%)} & \textbf{0.761(+5.20\%)} & 0.943(+0.29\%) & 0.809(-3.68\%) \\
ATM-GAD & \textbf{0.911(+4.28\%)} & 0.871(+0.94\%) & 0.944(+1.53\%) & 0.742(+2.61\%) & {\ul{0.962(+2.32\%)}} & {\ul{0.869(+3.54\%)}} \\ \hline
\end{tabular}%
}
\end{table*}

\subsection{Additional Results}
\noindent\textbf{Ablation studies.}
The proposed method, ATM-GAD, includes four key components: temporal motif extractor, adaptive $\delta$ selection, and two types of motif attention, as detailed in Section \ref{sec:methods}. To evaluate the effectiveness of each component, we conduct an ablation study using a progressive strategy. The results are presented in Table \ref{table::Abaltione} and additional results are presented in Appendix \ref{APP::Ablation Study}. Specifically, we commence from GCN that ignores any components. We first consider the temporal motif extractor, denoted as \textit{TM+$\delta_\mathrm{fixed}$}, then we replace the fixed window with our adaptive $\delta$ module, denoted as  \textit{TM+$\delta_\mathrm{ada}$}. The motif embeddings of all the nodes are computed as the average of the node embeddings. Then, we take into consideration our two types of attention mechanisms, denoted as \textit{TM+$\delta_\mathrm{ada}$+IntraA} and \textit{TM+$\delta_\mathrm{ada}$+InterA}, respectively. Finally, we add all the modules, which is the proposed ATM-GAD. From Table \ref{table::Abaltione}, we showcase that each component leads to a consistent improvement in both AUC and AUPRC in most cases. Interestingly, we find that two types of attention when coupled with each other achieve larger improvements compared to GCN, while \textit{TM+$\delta_\mathrm{ada}$+IntraA} or \textit{TM+$\delta_\mathrm{ada}$+InterA} alone fail to surpass GCN in some cases. 
 






\noindent\textbf{Parameter sensitivity.}
We further evaluate how changing the embedding dimension, number of GCN layers, and hidden dimension affects the performance. From Figure~\ref{fig::PC} (and additional results in Appendix~\ref{APP::Parameter sensitivity}), the results show that ATM-GAD 
remains robust across a wide range of embedding and hidden sizes. However, overly large hidden dimensions on smaller graphs (e.g., \textit{ETH-200}) can lead to overfitting. While 3-layer GCNs often offer a small advantage, the model’s performance degrades only slightly with fewer or more layers, demonstrating ATM-GAD’s stability.

\begin{figure}
    \centering
    \includegraphics[width=0.85\linewidth]{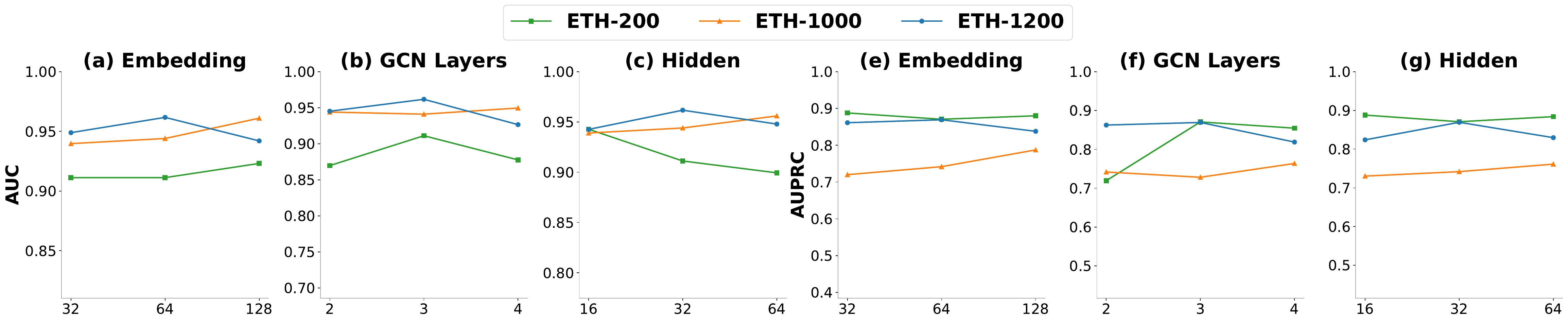}
    \captionsetup{skip=8pt}
    \caption{Parameter sensitivity studies of ATM-GAD.}
    \label{fig::PC}
\end{figure}

\section{Conclusion}\label{sec:conclusion}

In this paper, we introduced ATM-GAD, an adaptive GNN model for financial fraud detection leveraging temporal motifs to capture high-order, time-sensitive patterns in transaction networks. Our per-node motif approach ensures extracted patterns accurately reflect real transactions. Two specialized attention mechanisms capture relationships within motifs and integrate information across different structures. Experiments on four datasets demonstrate ATM-GAD outperforms state-of-the-art methods. 
 
\noindent\textbf{Limitations and future work.} 
While ATM-GAD demonstrates strong performance across multiple financial fraud detection scenarios, computational efficiency remains a challenge for large-scale applications. Future work includes enhancing computational efficiency and exploring heterophily in temporal motif construction.

\noindent\textbf{Broader impacts.} Our ATM-GAD improves financial fraud detection accuracy through temporal motifs, which has positive societal impact. However, there is minimal risk of overreliance on automation without human oversight, potentially causing erroneous decisions in complex cases.
\bibliographystyle{plain}
\bibliography{refs}






\end{document}